\documentclass[11pt]{article}

\usepackage[T1]{fontenc}
\usepackage[preprint]{acl}
\usepackage{times}
\usepackage{latexsym}

\usepackage{amsmath}
\usepackage{amssymb}
\usepackage{booktabs}
\usepackage{array}
\usepackage{tabularx}
\usepackage{longtable}
\usepackage{graphicx}
\usepackage{float}
\usepackage{algorithm}
\usepackage{algpseudocode}
\usepackage{placeins}
\usepackage{microtype}
\usepackage{xcolor}
\usepackage{colortbl}
\usepackage{listings}
\usepackage{etoolbox}
\usepackage{hyperref}
\usepackage{url}
\definecolor{codebg}{RGB}{247,247,247}
\definecolor{maskblue}{RGB}{215,232,250}
\definecolor{lossred}{RGB}{190,50,50}
\newcommand{\accl}{ACC\textsubscript{\textit{L}}}

\title{
    \textsc{DiffuPlan}: Editable Planning State for Diffusion Language Model Agents
}

\author{
  \textbf{Qisheng Zhou}\textsuperscript{1},
  \textbf{Zhen Xiong}\textsuperscript{2},
  \textbf{Qiaoyu Tan}\textsuperscript{1}\thanks{Corresponding author.}
\\
  \textsuperscript{1}New York University Shanghai,
  \textsuperscript{2}New York University
}
\date{}

\begin{document}

\maketitle

\begin{abstract}
Diffusion language models (dLLMs) are emerging as an alternative to autoregressive (AR) language models, yet dLLM agents largely inherit an interaction protocol designed for AR generation: once intermediate state is written, later observations can only be incorporated by appending more text. This is a poor match to a defining capability of dLLMs---reconstructing earlier text conditioned on context that appears both before and after it. We ask whether this capability can support a different agent-state paradigm in which intermediate planning state is directly revisable as evidence arrives. We introduce \textsc{DiffuPlan}, a diffusion search agent that maintains a single editable planning state and reconstructs only the affected steps when later evidence resolves an entity or changes the search route. To support variable-length revisions, we introduce Expansion-Partition Marginalization (EPM), which marginalizes over equivalent expansion paths. We further adapt diffusion trajectory RL to multi-turn search. Across four multi-hop QA benchmarks, \textsc{DiffuPlan} achieves 50.24 average F1, outperforming strong AR search agents. More importantly, under identical data and training, editable planning state improves the diffusion agent by 2.44 F1 versus only 0.47 for its AR source model. These results show that dLLMs need not imitate append-only AR agents: revising committed state is a distinct agentic advantage.\footnote{Code is available at \url{https://github.com/Jaasssoooonnnnn/DiffuPlan}.}
\end{abstract}

\section{Introduction}
\label{sec:introduction}

Language agents act over multiple turns, continually forming and revising intermediate hypotheses, subgoals, and plans as new observations arrive. Yet the textual state of most agents is effectively \emph{append-only}: once a model writes an intermediate decision, later evidence does not change that text; instead, the agent expresses the revision by generating additional text. This interaction pattern is natural for autoregressive (AR) language models, which generate a continuation conditioned on a fixed prefix. It is not obviously natural for diffusion language models (dLLMs).

dLLMs provide a different computational primitive. Rather than committing every token strictly left-to-right, masked and block-diffusion language models can reconstruct selected positions from surrounding context~\citep{nie2025llada,ye2025dream,arriola2025blockdiffusion,bie2025llada2}. This raises a basic question for agent design: \emph{should a diffusion agent inherit the same append-only state paradigm as an AR agent, or should its intermediate state itself remain editable?}

Existing dLLM agents largely follow the former design. They retain ReAct-style, append-only interaction histories~\citep{yao2023react} and use diffusion primarily to accelerate or restructure forward generation---for example, by issuing tool calls from partially denoised states or overlapping reasoning with tool execution~\citep{zhao2026dllmsearcher,sardi}. Under such an interaction, the distinctive reconstruction capability of dLLMs is not used to revise agent state across turns. In our controlled setting, this mismatch is visible empirically: with the same training data and post-training recipe, a search agent built on LLaDA2.0-mini reaches 47.80 average F1 under append-only plan revision, below 49.54 for Ling-mini-2.0~\citep{ling2025}, the AR model from which LLaDA2.0-mini was converted~\citep{bie2025llada2}.

Multi-hop search provides a natural testbed for studying this mismatch. During search, later evidence routinely changes state formed earlier: a retrieved passage can resolve an entity referenced by a pending subquestion, add constraints that make a query more specific, or invalidate the planned search route. Making this state explicit and revisable helps even without training: on our validation set, prompting the untrained LLaDA2.0-mini yields 26.53 F1 without a plan, 28.61 with an initial plan, and 29.26 when the plan is revised as evidence arrives. Existing agents can represent such changes, but typically by appending another rationale or a regenerated plan~\citep{d2plan}. Consequently, multiple versions of the same intermediate state may coexist in the interaction history. In contrast, a diffusion agent could maintain a single current state and revise the affected content \emph{where it was originally written}, using later evidence as conditioning context.

\begin{figure*}[!t]
    \centering
    \includegraphics[width=\textwidth]{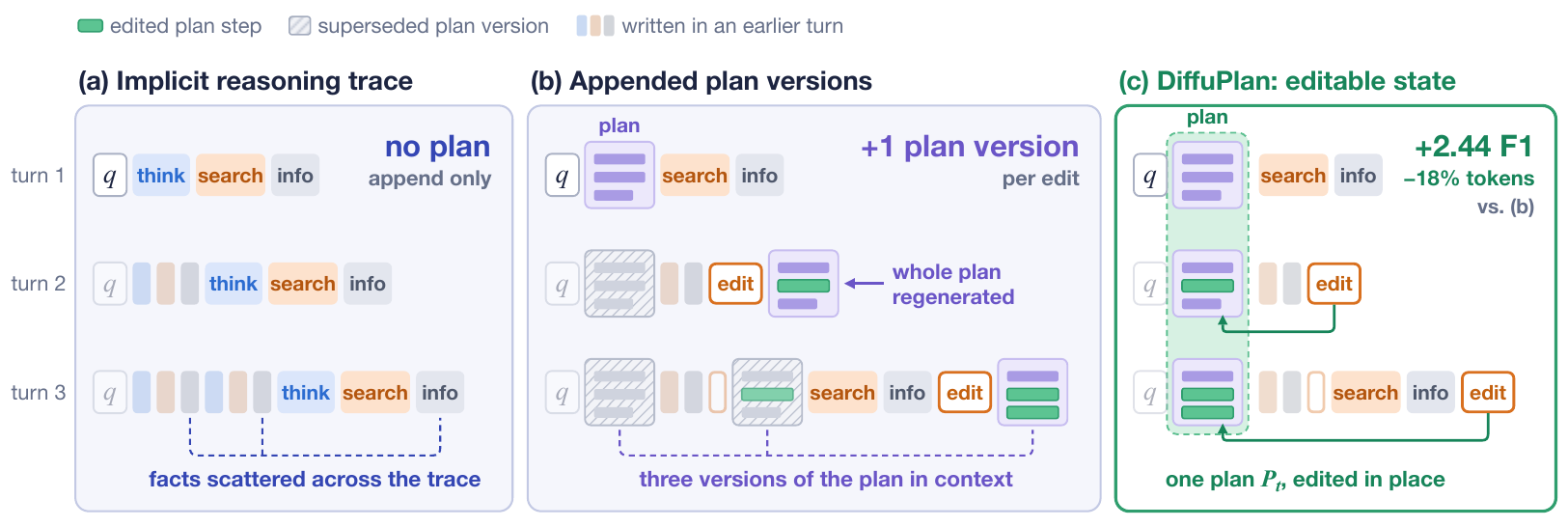}
    \caption{Three ways to represent evolving search state. (a) An implicit reasoning trace scatters intermediate state across appended thoughts and actions. (b) An explicit planner appends a regenerated plan after each revision, so superseded versions remain in the context. (c) \textsc{DiffuPlan} maintains one planning state $P_t$ and reconstructs only the affected steps in place using evidence retrieved later. Green marks revised content. The reported numbers compare \textsc{DiffuPlan} with an append-only diffusion agent trained with the same data and post-training recipe (Section~\ref{sec:state_transitions}).}
    \label{fig:overview}
\end{figure*}

We introduce \textsc{DiffuPlan}, a diffusion search agent with an \emph{editable planning state}. The plan is treated not merely as another piece of historical text, but as a persistent state variable that should reflect the agent's current search strategy. When later evidence resolves an intermediate answer or changes the route, the agent requests an edit and reconstructs only the affected plan span conditioned on the preserved plan and the subsequent interaction. This turns cross-turn state revision into a native reconstruction problem for the dLLM.

Making this idea trainable raises two central technical challenges. First, one shared model must support two different computation patterns: forward action generation should remain block-causal, whereas state revision must condition on later evidence. We therefore use a block-causal \emph{Action Mode} and a full-canvas \emph{Edit Mode} with shared parameters. Second, revisions are often variable-length. Dynamic expansion allows an editable mask to grow, but the same target replacement may be realized by many equally valid expansion paths. We introduce \emph{Expansion-Partition Marginalization} (EPM), which marginalizes over these paths and supervises the final edit rather than an arbitrary expansion trajectory. SFT teaches action generation and state reconstruction jointly, and we further optimize search decisions from end-task reward with \emph{Multi-turn Diffusion Search with Anchored Stabilization} (MDSAS), which extends TraceRL~\citep{tracerl} to multi-turn search and stabilizes training with a frozen SFT anchor.

The central empirical question is not merely whether editable plans improve QA, but whether they provide a \emph{diffusion-specific} advantage. Our controlled $2\times2$ comparison varies both backbone (diffusion vs.\ its AR source model) and state protocol (editable vs.\ append-only) under matched data and training. Editable state improves the dLLM by 2.44 F1, compared with only 0.47 F1 for the AR model, while reducing generated tokens for both. Moreover, the diffusion gain is mechanism-aligned: it is only 0.20 F1 when no edit is requested, but rises to 4.63 F1 with multiple updates and 7.12 F1 when replanning is required.

Our contributions are threefold:
\begin{itemize}\setlength{\itemsep}{1pt}\setlength{\parsep}{0pt}\setlength{\topsep}{2pt}
    \item We identify a mismatch between the append-only state paradigm inherited from AR agents and the cross-context reconstruction capability of dLLMs, and formulate \emph{editable planning state} as a diffusion-native alternative for multi-turn agents.
    \item We introduce \textsc{DiffuPlan}, which realizes cross-turn state revision with shared Action/Edit modes and learns variable-length edits with EPM, marginalizing over equivalent expansion paths.
    \item We show through matched backbone--protocol comparisons that editable state benefits the diffusion agent substantially more than its AR counterpart, with gains increasing as more revision is required, while \textsc{DiffuPlan} also achieves the best average F1 among the evaluated search agents.
\end{itemize}

\section{Related Work}
\label{sec:related_work}

\paragraph{Search agents and evolving intermediate state.}
Retrieval-augmented generation~\citep{lewis2020rag,gao2023ragsurvey} has developed into search agents that interleave reasoning with retrieval and decide what evidence to acquire next. IRCoT and ReAct alternate reasoning and search~\citep{trivedi2023ircot,yao2023react}; Search-R1 and R1-Searcher learn search behavior with reinforcement learning~\citep{jin2025searchr1,song2025r1searcher}; Search-o1 refines retrieved evidence during long-form reasoning~\citep{li2025searcho1}. A complementary line of work makes intermediate state more explicit. Plan$^{*}$RAG represents dependencies among retrieval steps~\citep{planstarrag}, D$^{2}$Plan revises subquestions and plans from retrieval feedback~\citep{d2plan}, and SLEUTH tracks facts, hypotheses, and open questions through successive structured states~\citep{sleuth}. These methods recognize that search state evolves, but their revisions are still serialized as newly generated text. \textsc{DiffuPlan} instead treats the current plan as a persistent state variable and directly reconstructs the affected earlier span after later evidence arrives.

\paragraph{Diffusion language models and text revision.}
Masked and block-diffusion language models generate by iterative denoising rather than strictly token-by-token continuation~\citep{nie2025llada,ye2025dream,arriola2025blockdiffusion,cheng2025sdar,bie2025llada2}. This makes reconstruction and editing natural operations. DreamOn introduces expansion and deletion states for variable-length infilling, while LLaDA2.1 revises previously committed tokens during block-diffusion generation~\citep{dreamon,llada21}. These works revise text \emph{within a single generation}. Our setting is different: an agent first commits a state, interacts with an external environment, receives new evidence in later turns, and then revises the earlier state conditioned on that evidence. We study this \emph{cross-turn} revision as an agent-state primitive.

\paragraph{Diffusion language models for agents and post-training.}
Recent work begins to adapt dLLMs to tool use and search. DLLM-Searcher combines SFT and preference optimization with P-ReAct, using partially denoised states to trigger retrieval early~\citep{zhao2026dllmsearcher}, while SARDI forms retrieval queries from tentative denoising states~\citep{sardi}. TraceRL optimizes denoising trajectories with reinforcement learning~\citep{tracerl}. These methods exploit diffusion during forward response generation, but the interaction history itself remains append-only. \textsc{DiffuPlan} is complementary: its central question is whether diffusion reconstruction can be used to revise agent state that was written in an earlier turn.

\section{DiffuPlan}
\label{sec:diffuplan}

\begin{figure*}[!t]
    \centering
    \includegraphics[width=\textwidth]{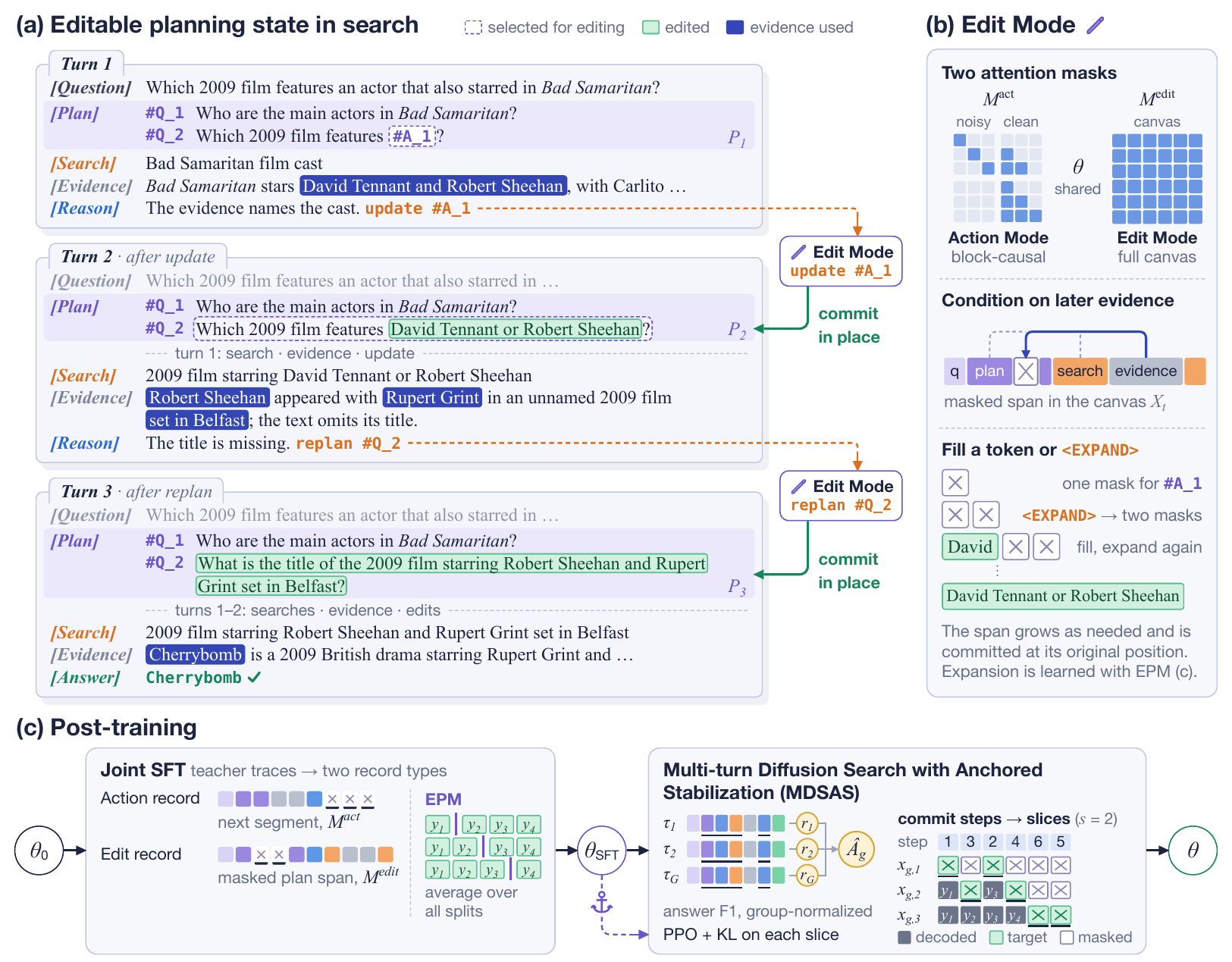}
    \caption{\textsc{DiffuPlan} runtime and training. (a) Action Mode searches and requests an update or replan; Edit Mode then reconstructs the selected earlier plan span using the evidence retrieved after it and commits the revision at the same position. (b) The shared dLLM uses block-causal attention $M^{\mathrm{act}}$ for forward action generation and full-canvas attention $M^{\mathrm{edit}}$ for cross-turn state reconstruction. (c) Joint SFT trains both views. Dynamic expansion supports variable-length edits, while EPM marginalizes over equivalent expansion paths. Multi-turn Diffusion Search with Anchored Stabilization (MDSAS) then optimizes multi-turn Action-Mode decisions while regularizing toward the frozen SFT policy.}
    \label{fig:diffuplan}
\end{figure*}

\textsc{DiffuPlan} treats planning as a mutable state-transition process. Let $H_t$ denote the interaction history available at turn $t$ and let $P_t$ denote the agent's current planning state, represented by a single plan embedded at a fixed location in that history. Standard append-only agents leave earlier plan text unchanged and express a revision by generating another version later in the sequence. \textsc{DiffuPlan} instead maintains the invariant that $P_t$ is the unique current plan: later evidence may trigger a state transition $P_t\rightarrow P_{t+1}$ that replaces only the affected span at its original position. The same dLLM implements both forward interaction and state reconstruction through two attention modes (Figure~\ref{fig:diffuplan}).

\subsection{Editable Planning State}
\label{sec:persistent_plan}

\paragraph{Planning state and actions.}
The initial plan decomposes a question into indexed subquestions \texttt{\#Q\_i}. A future step may refer to an unresolved intermediate answer through \texttt{\#A\_i}. At each turn, Action Mode generates reasoning followed by one of four terminal actions: \texttt{search} retrieves evidence, \texttt{update} binds a resolved intermediate answer into the plan, \texttt{replan} revises upcoming steps, and \texttt{answer} terminates the interaction. The two edit operations, \texttt{update} and \texttt{replan}, change the planning state rather than append a second plan.

Formally, given question $q$ and history $H_t$, Action Mode generates the next segment $z_t$ block by block under block-causal attention $M^{\mathrm{act}}$:
\begin{equation}
\begin{gathered}
z_t\sim\pi^{\mathrm{act}}_\theta(\cdot\mid q,H_t;M^{\mathrm{act}}),\\
(a_t,u_t)=\operatorname{terminal}(z_t),
\end{gathered}
\label{eq:action_mode}
\end{equation}
where $a_t\in\{\texttt{search},\texttt{update},\texttt{replan},\texttt{answer}\}$ and $u_t$ is its payload. A \texttt{search} appends a retrieved observation to the environment history. An \texttt{update} or \texttt{replan} instead hands control to Edit Mode before the next Action-Mode segment is generated.

\paragraph{Cross-turn state reconstruction.}
The key asymmetry is temporal: the plan span to be revised appears \emph{before} the evidence that motivates its revision. Controller $\Gamma$ maps an edit request $(a_t,u_t)$ to one or more editable spans in $P_t$. Edit Mode forms a full canvas $X_t$ containing the original plan, subsequent actions and observations, and the triggering edit request, then masks only the selected plan content. Under full-canvas attention $M^{\mathrm{edit}}$, the masked span can condition simultaneously on preserved plan context and on evidence that arrived later. The reconstructed text is committed at the original location, yielding the new state $P_{t+1}$ while leaving unselected state and the surrounding interaction unchanged.

For a replan that opens the suffix beginning at a selected step, write
$P_t=P_t^{\mathrm{keep}}\oplus P_t^{\mathrm{open}}$. Edit Mode reconstructs
$\widehat P_t^{\mathrm{open}}$ and commits
\begin{equation}
P_{t+1}=P_t^{\mathrm{keep}}\oplus\widehat P_t^{\mathrm{open}}.
\end{equation}
An \texttt{update} is a localized special case: the selected answer marker is reconstructed with its resolved value, which is then propagated to identical unresolved references in the plan. Appendix~\ref{app:runtime} gives the complete execution loop and Appendix~\ref{app:controller} specifies localization and commit semantics.

\paragraph{Dynamic expansion.}
State revisions need not preserve length. A placeholder may expand into a multi-token entity name, and replanning may replace a short subquestion with a more specific or longer continuation. We therefore allow an editable mask either to emit a token or to emit
$\chi_{\mathrm{exp}}=\langle\mathrm{EXPAND}\rangle$, which replaces that mask with two masks carrying the same owner label. Repeated fill-or-expand operations support length-preserving and length-growing edits without reserving a fixed-size canvas.

\subsection{Learning Variable-Length State Revision}
\label{sec:editable_sft}

Joint SFT teaches two complementary behaviors from the same teacher trajectories. An \emph{action record} supervises the next Action-Mode segment given the current interaction history. An \emph{edit record} supervises the desired replacement given a masked earlier plan, the later evidence, and the triggering edit request. The two record types share parameters but use the attention pattern of their corresponding runtime mode.

\paragraph{Why expansion creates ambiguous supervision.}
Dynamic expansion introduces a latent-path problem. Suppose two active masks must reconstruct four target tokens $(y_1,y_2,y_3,y_4)$. The target edit is unique, but the four tokens can be allocated to the two masks as $(1,3)$, $(2,2)$, or $(3,1)$. These allocations imply different immediate labels---for example, $(y_1,\chi_{\mathrm{exp}})$ versus $(\chi_{\mathrm{exp}},\chi_{\mathrm{exp}})$---even though they can lead to exactly the same final replacement. Standard single-path cross-entropy chooses one allocation as the target and therefore penalizes other valid expansion decisions.

\paragraph{Expansion-Partition Marginalization.}
EPM removes this arbitrary choice by marginalizing over all valid allocations. Let $\mathcal U_e$ be the contiguous unresolved components of edit record $e$. Component $c$ has target sequence $y^{(c)}$ of length $n_c$, $m_c$ active masks, and $\mathcal P_c$, the set of positive $m_c$-part compositions of $n_c$. For a composition $\boldsymbol\kappa$, mask $i$ is supervised with the corresponding target token when $\kappa_i=1$ and with $\chi_{\mathrm{exp}}$ otherwise. Let $\widetilde X_{e,c}$ be the full canvas at the current editing step, in which some target tokens of component $c$ may already be resolved, and let $i$ index its active masks. The edit objective is
\begin{equation}
\begin{aligned}
\ell_{\mathrm{edit}}(e)
&=-\sum_{c\in\mathcal U_e}
\log\!\Biggl[
\frac{1}{|\mathcal P_c|}
\sum_{\boldsymbol\kappa\in\mathcal P_c}
\varphi_c(\boldsymbol\kappa)
\Biggr],\\
\varphi_c(\boldsymbol\kappa)
&=\prod_{i=1}^{m_c}
p_\theta\!\left(
\lambda_i(\boldsymbol\kappa)
\mid \widetilde X_{e,c},i;M^{\mathrm{edit}}
\right).
\end{aligned}
\label{eq:sft_objective}
\end{equation}
Thus EPM supervises the \emph{replacement} rather than an arbitrary expansion trajectory. Appendix~\ref{app:epm} gives a complete example and an $O(mn)$ dynamic program for the marginal.

\paragraph{Joint SFT.}
The Action-Mode objective $\ell_{\mathrm{act}}$ is the standard denoising loss over corrupted positions in the newly generated segment. We optimize action and edit records jointly:
\begin{equation}
\mathcal L_{\mathrm{SFT}}=\frac{1}{|\mathcal D_{\mathrm{mix}}|}
\Bigl[\sum_{r\in\mathcal D_{\mathrm{mix}}^{\mathrm{act}}}\ell_{\mathrm{act}}(r)
+\sum_{e\in\mathcal D_{\mathrm{mix}}^{\mathrm{edit}}}\ell_{\mathrm{edit}}(e)\Bigr],
\end{equation}
where the training multiset $\mathcal D_{\mathrm{mix}}=\mathcal D_{\mathrm{mix}}^{\mathrm{act}}\uplus\mathcal D_{\mathrm{mix}}^{\mathrm{edit}}$ contains action and edit records with the multiplicities given in Appendix~\ref{app:training}. SFT therefore learns \emph{how} to produce executable actions and \emph{how} to realize a requested state revision.

\begin{table*}[t]
\caption{End-to-end QA performance (\%) on four multi-hop benchmarks. Each dataset reports token F1 and LLM-as-Judge accuracy (\accl); Avg.\ is the unweighted mean across the four datasets. Bold marks the best score in each column.}
\label{tab:system_comparison}
\centering
\footnotesize
\setlength{\tabcolsep}{5pt}
\begin{tabular}{l*{10}{c}}
\toprule
& \multicolumn{2}{c}{HotpotQA} & \multicolumn{2}{c}{2Wiki} & \multicolumn{2}{c}{MuSiQue} & \multicolumn{2}{c}{Bamboogle} & \multicolumn{2}{c}{Avg.} \\
\cmidrule(lr){2-3}\cmidrule(lr){4-5}\cmidrule(lr){6-7}\cmidrule(lr){8-9}\cmidrule(lr){10-11}
\textbf{System} & F1 & \accl & F1 & \accl & F1 & \accl & F1 & \accl & F1 & \accl \\
\midrule
\rowcolor{black!8}
\multicolumn{11}{c}{\textit{Inference-only methods}} \\
Direct & 25.72 & 29.20 & 29.04 & 28.80 & 11.26 & 15.60 & 21.51 & 20.80 & 21.88 & 23.60 \\
CoT & 21.95 & 26.40 & 32.55 & 31.40 & 11.70 & 11.20 & 40.88 & 40.80 & 26.77 & 27.45 \\
Standard RAG & 39.30 & 46.40 & 31.96 & 30.60 & 10.60 & 11.40 & 25.06 & 29.60 & 26.73 & 29.50 \\
SuRe & 37.68 & 45.60 & 26.90 & 27.60 & 12.15 & 12.40 & 29.44 & 35.20 & 26.54 & 30.20 \\
Selective-Context & 35.38 & 42.80 & 28.82 & 28.40 & 11.15 & 11.40 & 27.09 & 32.80 & 25.61 & 28.85 \\
Iter-RetGen & 42.98 & 52.00 & 28.66 & 29.80 & 13.38 & 14.80 & 37.56 & 43.20 & 30.65 & 34.95 \\
IRCoT & 46.45 & 54.40 & 37.56 & 37.60 & 18.68 & 19.60 & 36.83 & 38.40 & 34.88 & 37.50 \\
\midrule
\rowcolor{black!8}
\multicolumn{11}{c}{\textit{Autoregressive agents}} \\
Search-o1 & 43.29 & 55.40 & 36.90 & 37.80 & 21.78 & 27.00 & 57.16 & 64.80 & 39.78 & 46.25 \\
Search-R1 & 47.26 & 52.40 & 27.90 & 28.40 & 18.22 & 20.60 & 47.93 & 49.60 & 35.33 & 37.75 \\
R1-Searcher & 53.94 & 62.80 & 52.50 & 52.40 & 30.88 & 32.00 & 57.53 & 63.20 & 48.71 & 52.60 \\
\midrule
\rowcolor{black!8}
\multicolumn{11}{c}{\textit{Diffusion agents}} \\
DLLM-Searcher & 7.18 & 48.60 & 9.68 & 36.20 & 4.91 & 18.60 & 7.25 & 43.20 & 7.25 & 36.65 \\
\textsc{DiffuPlan} & \textbf{55.34} & \textbf{65.00} & \textbf{55.03} & \textbf{56.00} & \textbf{31.72} & \textbf{34.60} & \textbf{58.86} & \textbf{65.60} & \textbf{50.24} & \textbf{55.30} \\
\bottomrule
\end{tabular}\end{table*}

\subsection{Optimizing Multi-Turn Search Decisions}
\label{sec:action_optimization}

SFT imitates teacher actions along teacher trajectories, but the agent must also learn \emph{when} to search, update, replan, or answer under its own rollouts. We therefore further optimize these decisions with MDSAS, which adapts TraceRL~\citep{tracerl} from single-response diffusion generation to multi-turn search. Importantly, this stage optimizes Action-Mode decisions; Edit-Mode reconstruction remains supervised by SFT.

For each question with answer aliases $\mathcal Y^\star$, we sample $G$ complete trajectories $\{\tau_g\}_{g=1}^G$. The terminal reward is answer F1 multiplied by a format-validity gate and is normalized within each question group to obtain a trajectory-level advantage. As in TraceRL, rollout records the denoising step at which each Action-Mode token is committed. We group every few commit steps of a block into a slice $x_{g,k}$, and evaluate the tokens committed in each slice given the tokens committed before it. The resulting clipped policy-gradient objective is denoted $\mathcal L_{\mathrm{RL}}^{\mathrm{act}}$; full details are given in Appendix~\ref{app:ppo}.

\paragraph{Frozen-reference stabilization.}
Multi-turn agent RL introduces a failure mode not addressed by a moving old-policy reference: once all rollouts in a group become structurally invalid, their rewards have no variance and the policy receives no signal that restores the executable action format. As in KL-regularized RL fine-tuning~\citep{ouyang2022instructgpt,shao2024deepseekmath}, we therefore regularize every rollout slice toward the frozen SFT policy $\theta_{\mathrm{SFT}}$. Let
\begin{equation}
u_{g,k,j}
=
\log p_{\theta_{\mathrm{SFT}},g,k,j}
-
\log p_{\theta,g,k,j},
\end{equation}
and $h(u)=\exp(u)-1-u$. The final objective is
\begin{equation}
\begin{aligned}
\mathcal L_{\mathrm{anchor}}
&=
\mathbb E_{\Omega_{\mathrm{all}}}[h(u_{g,k,j})],\\
\mathcal L_{\mathrm{MDSAS}}
&=
\mathcal L_{\mathrm{RL}}^{\mathrm{act}}
+
\lambda_{\mathrm{SFT}}\mathcal L_{\mathrm{anchor}}.
\end{aligned}
\label{eq:sft_anchor}
\end{equation}
The anchor is a training stabilization mechanism rather than the core state-revision contribution; Section~\ref{sec:analysis_action_optimization} quantifies its effect.

\section{Experiments}
\label{sec:experiments}

Our experiments answer four questions:
\textbf{RQ1:} Does \textsc{DiffuPlan} provide competitive end-to-end multi-hop search performance?
\textbf{RQ2:} Does editable planning state benefit a diffusion backbone more than an otherwise matched AR backbone?
\textbf{RQ3:} Does the benefit increase on examples that require more state revision, as predicted by our hypothesis?
\textbf{RQ4:} Are EPM, dynamic expansion, and the frozen SFT anchor necessary for learning and optimizing the editable-state agent?

\subsection{Experimental Setup}
\label{sec:setup}

\paragraph{Benchmarks.}
We evaluate on four open-domain multi-hop QA benchmarks in which later evidence can change earlier intermediate state: 500 sampled questions each from HotpotQA~\citep{yang2018hotpotqa}, 2WikiMultiHopQA (2Wiki)~\citep{ho2020twowiki}, and MuSiQue~\citep{trivedi2022musique}, plus all 125 Bamboogle~\citep{press2023bamboogle} questions. Teacher trajectories are collected only from the HotpotQA and 2Wiki training splits; MuSiQue and Bamboogle therefore test transfer beyond the teacher-trajectory domains.

\paragraph{Baselines and controlled protocol.}
Inference-only baselines include Direct, CoT~\citep{kojima2022zeroshot}, Standard RAG~\citep{lewis2020rag}, SuRe~\citep{kim2024sure}, Selective-Context~\citep{li2023selectivecontext}, Iter-RetGen~\citep{shao2023iterretgen}, and IRCoT~\citep{trivedi2023ircot}, all using the unmodified LLaDA2.0-mini checkpoint. We compare against AR search agents Search-o1~\citep{li2025searcho1}, Search-R1~\citep{jin2025searchr1}, and R1-Searcher~\citep{song2025r1searcher}, using their released systems. For DLLM-Searcher~\citep{zhao2026dllmsearcher}, we migrate its complete post-training pipeline, including SFT and Variance-Reduced Preference Optimization (VRPO), to LLaDA2.0-mini and use its P-ReAct inference procedure. All retrieval-capable systems query the same frozen E5/Wiki18 index.

To test the central hypothesis independently of overall system differences, we additionally construct a matched $2\times2$ comparison: \{editable, append-only\} planning state on \{LLaDA2.0-mini, Ling-mini-2.0\}. Ling-mini-2.0~\citep{ling2025} is the AR model from which LLaDA2.0-mini was converted~\citep{bie2025llada2}, with the same architecture, parameter count, and tokenizer; the four agents use the same teacher trajectories, SFT schedule, MDSAS recipe, retriever, retrieval budget, and checkpoint-selection protocol. This comparison isolates whether the state protocol interacts specifically with the diffusion backbone.

\paragraph{Metrics.}
Token F1 is our primary answer metric, following R1-Searcher~\citep{song2025r1searcher}. We additionally report LLM-as-Judge accuracy (\accl) using Qwen3.5-9B~\citep{qwen35} to account for semantically correct answers that differ lexically from the reference. Appendix~\ref{app:answer_output} reports EM and SubEM. We also measure model-generated tokens and format validity where relevant.

\paragraph{Training and inference.}
Qwen3.6-27B generates teacher trajectories used for joint Action/Edit SFT of LLaDA2.0-mini. MDSAS starts from the epoch-one SFT checkpoint, which also provides the frozen anchor. RL checkpoints are selected on a disjoint 256-question HotpotQA--2Wiki validation set. At inference, \textsc{DiffuPlan} uses greedy decoding with at most five top-3 retrievals. Additional hyperparameters and data construction details are in Appendices~\ref{app:teacher_data} and~\ref{app:training}.

\subsection{End-to-End Search Performance}
\label{sec:system_comparison}

Table~\ref{tab:system_comparison} evaluates overall answer quality. Under the semantic judge, \textsc{DiffuPlan} reaches 55.30 average \accl{}, exceeding R1-Searcher by 2.70 points. Token F1 gives the same ranking: \textsc{DiffuPlan} obtains the highest average F1, 50.24, exceeding R1-Searcher by 1.53 points. It achieves the best score on all four datasets under both metrics. Gains on MuSiQue and Bamboogle show that the learned behavior transfers beyond the HotpotQA and 2Wiki domains used to construct teacher trajectories.

DLLM-Searcher produces substantially more verbose answer strings, which makes token F1 unusually harsh for that system. Under SubEM, the containment-style metric used in its evaluation setting, \textsc{DiffuPlan} still leads, 48.45 versus 36.25 (Appendix~\ref{app:answer_output}). These results establish that editable-state diffusion search is competitive end-to-end; the following analyses test \emph{why}.

\begin{table}[t]
\caption{Editable versus append-only planning state, trained with the same data and recipe. Token F1 and \accl{} (\%) are averaged over the four datasets, and Tokens is model-generated tokens per question.}
\label{tab:backbone_protocol}
\centering
\footnotesize
\setlength{\tabcolsep}{6pt}
\begin{tabular}{lccc}
\toprule
\textbf{Planning state} & F1 & \accl & Tokens \\
\midrule
\multicolumn{4}{l}{\textit{dLLM (LLaDA2.0-mini)}} \\
Editable & \textbf{50.24} & \textbf{55.30} & \textbf{338} \\
Append & 47.80 & 52.30 & 411 \\
\midrule
\multicolumn{4}{l}{\textit{AR LLM (Ling-mini-2.0)}} \\
Editable & 50.01 & 53.70 & 357 \\
Append & 49.54 & 52.60 & 425 \\
\bottomrule
\end{tabular}
\end{table}

\section{Testing the Editable-State Hypothesis}
\label{sec:analysis}

\subsection{Does Editable State Specifically Benefit Diffusion?}
\label{sec:state_transitions}

Our central hypothesis predicts an interaction effect: keeping one current plan should reduce redundant generation for either backbone, but direct reconstruction of earlier state should provide a larger accuracy benefit to the dLLM than to the AR model. Table~\ref{tab:backbone_protocol} supports this prediction. On LLaDA2.0-mini, editable state improves average F1 from 47.80 to 50.24 (+2.44) and reduces model-generated tokens from 411 to 338 (18\%). On Ling-mini-2.0, the same state protocol changes F1 from 49.54 to 50.01 (+0.47), despite a comparable token reduction from 425 to 357 (16\%). The corresponding \accl{} gains are +3.00 for the diffusion model and +1.10 for the AR model.

This comparison is important because both editable agents ultimately maintain the same current plan. The AR agent generates replacement text left-to-right and the controller writes it into the plan; the diffusion agent reconstructs the selected earlier span directly under full-canvas conditioning. Thus the much larger dLLM gain cannot be explained by retaining a single plan or by token savings alone. It is consistent with the hypothesis that bidirectional reconstruction supplies a better inductive bias for revising previously committed state from later observations. Although Ling-mini-2.0 scores slightly higher than LLaDA2.0-mini on general benchmarks~\citep{bie2025llada2}, \textsc{DiffuPlan} is the most accurate and least costly of the four agents. Per-dataset results are reported in Appendix~\ref{app:ar_counterpart}.

\begin{figure}[t]
    \centering
    \includegraphics[width=0.95\columnwidth]{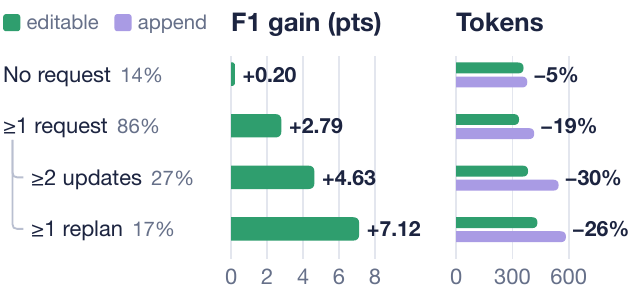}
    \caption{Editable versus append-only planning state on the 1{,}625 test questions, grouped by the append agent's edit requests (the last two groups are subsets of the second). The F1 gain grows with the number of state edits, and the editable agent generates fewer tokens in every group. Full numbers are in Table~\ref{tab:edit_usage_paired}.}
    \label{fig:edit_gain}
\end{figure}

\subsection{Does the Gain Track the Need for Revision?}
\label{sec:update_scaling}

A second prediction is that editable state should matter little when no state revision is needed and matter increasingly as revision becomes more substantial. Figure~\ref{fig:edit_gain} groups all 1{,}625 test questions by the \emph{append agent's} own edit requests, so the grouping is independent of \textsc{DiffuPlan}. The pattern is monotonic: editable state changes F1 by only +0.20 on questions with no edit request, by +2.79 when at least one edit is requested, by +4.63 when at least two updates are requested, and by +7.12 when replanning is requested. The editable agent also generates 5--30\% fewer tokens in every group. Overall, \textsc{DiffuPlan} edits its state on 83\% of test questions (Appendix~\ref{app:edit_usage}).

The accuracy trend is more informative than the efficiency gain alone. If the improvement came mainly from shorter outputs or from maintaining an explicit plan, we would not expect it to increase so sharply with the amount of required revision. Instead, the gain is concentrated precisely where later evidence most strongly changes earlier planning state.

\subsection{Are EPM and Expansion Necessary?}
\label{sec:epm_ablation}

\begin{figure}[!t]
    \centering
    \includegraphics[width=0.95\columnwidth]{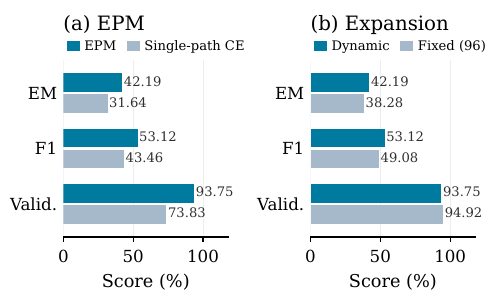}
    \caption{(a) EPM versus single-path cross-entropy (CE) and (b) dynamic expansion versus a fixed 96-token canvas, at SFT epoch~1 on the 256 validation questions under the same interaction budget. Valid.\ is format validity.}
    \label{fig:edit_learning}
\end{figure}

\paragraph{Marginalizing equivalent expansion paths.}
Figure~\ref{fig:edit_learning}(a) compares EPM with single-path cross-entropy while keeping dynamic expansion fixed. EPM improves validation EM from 31.64 to 42.19, F1 from 43.46 to 53.12, and format validity from 73.83\% to 93.75\%. The large gap supports the motivation for EPM: supervision should identify the desired replacement, not an arbitrary latent allocation of that replacement to expandable masks.

\paragraph{Allocating edit capacity dynamically.}
Figure~\ref{fig:edit_learning}(b) compares dynamic expansion with a fixed 96-token edit canvas. Dynamic expansion raises F1 from 49.08 to 53.12 while preserving comparable format validity. This is not a rare corner case: 90.7\% of executed edits on the test set invoke expansion (Appendix~\ref{app:edit_usage}). Variable-length state revision is therefore central to the practical protocol rather than an occasional fallback.

\subsection{Optimizing the Agent without Losing the Protocol}
\label{sec:analysis_action_optimization}

With its frozen SFT anchor, MDSAS improves the SFT initialization from 53.12 to 58.98 validation F1. The anchor is essential for stable optimization: without it, hard-invalid rollouts rise to 99.8\% by step 50 and validation F1 collapses to 2.08; with it, the hard-invalid fraction stays below 10\% over the same interval (Appendix~\ref{app:anchor_curves}). The best unanchored checkpoint reaches 55.81 F1, so the anchor improves not only late-stage stability but also the selected policy.

These results clarify the roles of the training components. SFT learns the editable-state protocol and the reconstruction operation; EPM makes variable-length revision learnable; and MDSAS improves multi-turn action selection, while its frozen SFT anchor keeps RL from destroying the executable protocol.

\section{Conclusion}
\label{sec:conclusion}

Current dLLM agents largely inherit an append-only interaction paradigm from autoregressive models, even though dLLMs can reconstruct earlier text from later context. \textsc{DiffuPlan} turns this unused capability into an agent-state primitive by maintaining one editable planning state and revising selected earlier spans as new evidence arrives. A shared Action/Edit model realizes forward acting and cross-turn reconstruction, while EPM learns variable-length edits without committing supervision to an arbitrary expansion path. Controlled experiments show that editable state improves the diffusion agent far more than its AR counterpart and that the gain increases with the amount of revision required. Together with strong end-to-end QA performance, these results suggest a broader design principle: diffusion agents need not simply reproduce AR-style interaction; they can exploit reconstruction to maintain state that remains revisable throughout interaction.

\section*{Limitations}

Our study uses one diffusion backbone, LLaDA2.0-mini, and its AR source model, Ling-mini-2.0. Multi-turn RL over denoising trajectories requires repeated rollouts with retrieval at every turn, which limited the scale of our experiments; whether the advantage of editable state holds for larger dLLMs or other diffusion families remains untested. We evaluate only multi-hop open-domain QA over a fixed Wikipedia corpus, where plans are short and structured as indexed subquestions. Longer-horizon agents, such as coding agents or research agents with live web search, hold less structured state, and our marker-based controller would need to generalize to it. MDSAS optimizes when to edit but not the reconstruction itself, which stays as learned in SFT. Finally, SFT relies on trajectories from a larger teacher, Qwen3.6-27B, on two training domains, so the protocol it learns depends on the teacher's search and revision behavior.

\FloatBarrier
\bibliography{references}

\clearpage
\appendix

\section{Detailed Runtime}
\label{app:runtime}

Algorithm~\ref{alg:diffuplan_runtime} gives the complete execution loop, including the edit handoff between two Action-Mode segments. Each interaction produces one Action-Mode segment under the following grammar:
\begin{lstlisting}[caption={Action grammar. The environment, never the policy, emits \texttt{<information>}.},label={lst:action-grammar}]
initial_segment ::= <plan> plan_steps </plan>
                    later_segment
later_segment   ::= <think> rationale </think>
                    terminal_action
terminal_action ::=
      <search> query </search>
    | <update> answer_markers </update>
    | <replan> step_markers </replan>
    | <answer> short_answer </answer>
answer_markers  ::= #A_i ("," #A_j)*
step_markers    ::= #Q_i ("," #Q_j)*
\end{lstlisting}
The grammar places a plan only in the initial segment and one terminal action after a non-empty \texttt{think}. Terminal answers must be concise. The runtime executes the first complete terminal action and requests a direct answer if none is available or an interaction budget is exhausted.

\begin{algorithm*}[t]
\caption{Closed-loop search with committed plan editing}
\label{alg:diffuplan_runtime}
\footnotesize
\begin{algorithmic}[1]
\Require question $q$, controller $\Gamma$, retriever $\mathcal R$, turn budget $B_{\mathrm{turn}}$, retrieval budget $K_{\mathrm{ret}}$
\State $H\gets\varnothing$; $n_{\mathrm{ret}}\gets0$
\For{$t=0,\ldots,B_{\mathrm{turn}}-1$}
    \State $z_t\gets\Call{DenoiseAction}{q,H;M^{\mathrm{act}}}$
    \State $(a_t,u_t)\gets\Call{FirstCompleteTerminal}{z_t}$
    \If{no complete terminal action exists} \State \Return \Call{DirectAnswer}{$q,H$} \EndIf
    \State $H\gets H\oplus z_t$ \Comment{the first valid segment contains $P_0$}
    \If{$a_t=\texttt{answer}$} \State \Return $u_t$
    \ElsIf{$a_t=\texttt{search}$}
        \If{$n_{\mathrm{ret}}=K_{\mathrm{ret}}$} \State \Return \Call{DirectAnswer}{$q,H$} \EndIf
        \State $H\gets H\oplus\Call{Information}{\mathcal R(u_t)}$; $n_{\mathrm{ret}}\gets n_{\mathrm{ret}}+1$
    \Else \Comment{edit handoff; Appendix~\ref{app:controller} gives the controller rules}
        \State $P_t\gets\Call{LatestPlan}{H}$; $S_t\gets\Gamma(P_t,a_t,u_t)$
        \State Form the full canvas $X_t$ from $q$ and $H$, including evidence after $P_t$, and mask the spans $S_t$
        \State Reconstruct the masked spans under $M^{\mathrm{edit}}$ by filling or expanding their masks
        \If{$a_t=\texttt{update}$} \State propagate each decoded value to all identical unresolved plan references \EndIf
        \State Replace $P_t$ by $P_{t+1}$ at its original position in $H$, preserving the surrounding history
    \EndIf
\EndFor
\State \Return \Call{DirectAnswer}{$q,H$}
\end{algorithmic}
\end{algorithm*}

\section{Controller Implementation}
\label{app:controller}

\subsection{Localize}

The policy selects edit targets through markers; the controller localizes them without inferring which steps are solved. An \texttt{update} selects the first occurrence of each requested \texttt{\#A\_i} present in the plan. A \texttt{replan} with one matching \texttt{\#Q\_i} selects the suffix starting at that step; with multiple matching markers, it selects only the listed step spans. Unselected text is preserved.

\subsection{Commit and history semantics}

For an update, each selected answer marker is replaced by one logical mask. For a replan, all tokens overlapping the selected step spans are masked. The editor reconstructs these positions from the full interaction canvas, including subsequent actions and retrieved evidence. Commit propagates each decoded update value to all identical unresolved references and writes the revised plan back at its original position in the history.

The shared denoiser $\mathcal D_\theta$ reconstructs the masked spans of canvas $X_t$. The complete transition formalizes span selection, reconstruction, and commit as
\begin{equation}
\begin{aligned}
    S_t&=\Gamma(P_t,a_t,u_t),\\
    (\widetilde P_t,\nu_t,\xi_t)
      &=\mathcal D_\theta(\operatorname{Mask}(X_t,S_t);M^{\mathrm{edit}}),\\
    K_t&=\operatorname{Close}(P_t,S_t,\nu_t,a_t),\\
    P_{t+1}&=\operatorname{Commit}(P_t,\widetilde P_t,K_t,\nu_t).
\end{aligned}
\label{eq:edit_transition}
\end{equation}
Here $\widetilde P_t$ is the candidate plan before reference propagation. The map $\nu_t$ contains each selected update marker's decoded value and is empty for replan. The commit closure $K_t$ collects the spans changed by direct editing or reference propagation. It equals the selected replan spans; for an update, it additionally includes every identical marker occurrence reached by propagation. Commit preserves all other plan content. The denoiser returns status $\xi_t\in\{\texttt{complete},\texttt{stalled},\texttt{max\_steps}\}$. Missing plans or requests with no matching spans leave the history unchanged.

Each editable mask carries an owner label $\omega$. Expansion replaces it with two masks bearing the same label, keeping new positions attached to the originating span. Commit preserves all unselected plan content in order, including when expansion shifts subsequent token positions.

\section{Action/Edit Serialization}
\label{app:serialization}

Action Mode and Edit Mode are two views of one shared checkpoint. For Action Mode, define the history tokens $h_t=\operatorname{Tok}(\operatorname{Prompt}(q)\Vert H_t)$, target tokens $y_t=\operatorname{Tok}(z_t)$, and clean sequence $x_0=h_t\Vert y_t$, where $z_t$ is the complete segment from Equation~\ref{eq:action_mode}. The first target contains \texttt{plan+think+search}; later targets contain \texttt{think} followed by one terminal action. Left truncation preserves $y_t$.

\subsection{Edit Mode full canvas}

The edit input is not an isolated plan. The triggering \texttt{update}/\texttt{replan} action has already been appended, and the original plan appears earlier in the same canvas:
\begin{lstlisting}[caption={Full-canvas Edit-Mode template.},label={lst:edit-template}]
X_t = <role>SYSTEM</role> ... <|role_end|>
  <role>HUMAN</role> ... Question: {q}
  <|role_end|>
  <role>ASSISTANT</role>
  <plan>
  {plan_left}{editable_span}{plan_right}
  </plan>
  {subsequent actions and
   <information> observations}
  <think>{edit rationale}</think>
  <update|replan>{selected markers}
  </update|replan>
\end{lstlisting}
For the latest plan span $[b_t,e_t)$, let $\mathcal G_t=(c_1,\ldots,c_d)$ be its ordered replacement groups. Each group renders as
\begin{equation}
\begin{aligned}
Z_t^{(c)}&=
\begin{cases}
y^{(c)}_{u_c:v_c},&c\text{ resolved},\\
\langle\mathrm{MASK}_{\omega_c}\rangle,&c\text{ unresolved},
\end{cases}\\
Z_t&=\operatorname{ReplaceGroups}\!\left(\operatorname{Tok}(P_t),\right.\\
&\qquad\left.\{c\mapsto Z_t^{(c)}\}_{c\in\mathcal G_t}\right).
\end{aligned}
\end{equation}
The full-canvas serialization is
\begin{equation}
x_t^{\mathrm{edit}}=\operatorname{Tok}(X_t[:b_t])\Vert Z_t\Vert
\operatorname{Tok}(X_t[e_t:]).
\end{equation}
Here $[u_c,v_c)$ are group $c$'s resolved target-token bounds and $\omega_c$ is its controller owner. An unresolved group whose target length is one is labelled with that token; a longer group is labelled with $\chi_{\mathrm{exp}}$ (\texttt{<EXPAND>}) and trained with EPM. SFT constructs partially resolved canvases from the known replacement, whereas a runtime update begins with one logical mask per selected answer marker. They share full-canvas conditioning and expansion semantics, but need not start from the identical corruption state.

\subsection{Attention masks}

For Action-Mode SFT, let $x_\sigma=\operatorname{Corrupt}_\sigma(x_0;y_t)$ mask only positions in the new target segment. The trainer receives the length-$2N$ sequence $[x_\sigma;x_0]$, where $\sigma$ is the sampled corruption rate, and both lanes reuse position IDs. Let $\operatorname{lane}(i)\in\{\sigma,0\}$ denote the noisy or clean lane and $\operatorname{blk}(i)=\lfloor(i\bmod N)/L_{\mathrm{blk}}\rfloor$, where $N$ is the single-lane sequence length and $L_{\mathrm{blk}}=32$ is the diffusion block length. Ignoring padding,
\begin{equation}
\begin{gathered}
M^{\mathrm{act}}_{ij}=1\Longleftrightarrow\\
\begin{cases}
\begin{aligned}
&\operatorname{lane}(i)=\operatorname{lane}(j),\\
&\operatorname{blk}(i)=\operatorname{blk}(j),\\
&\text{same-lane current block},
\end{aligned}\\
\begin{aligned}
&\operatorname{lane}(i)=\sigma,\ \operatorname{lane}(j)=0,\\
&\operatorname{blk}(j)<\operatorname{blk}(i),\\
&\text{earlier clean blocks},
\end{aligned}\\
\begin{aligned}
&\operatorname{lane}(i)=0,\ \operatorname{lane}(j)=0,\\
&\operatorname{blk}(j)\le\operatorname{blk}(i),\\
&\text{clean block-causal lane}.
\end{aligned}
\end{cases}
\end{gathered}
\label{eq:mact}
\end{equation}
Table~\ref{tab:mact-blocks} shows the resulting block visibility pattern.
\begin{table}[tbp]
\caption{Block-level Action mask; 1 marks a visible key block.}
\label{tab:mact-blocks}
\centering
\footnotesize
\setlength{\tabcolsep}{5pt}
\renewcommand{\arraystretch}{1.25}
\begin{tabular}{@{}c|ccc|ccc@{}}
Query$\,\backslash\,$key & $x_\sigma^0$ & $x_\sigma^1$ & $x_\sigma^2$ & $x_0^0$ & $x_0^1$ & $x_0^2$ \\
\hline
$x_\sigma^0$ & 1 & $\cdot$ & $\cdot$ & $\cdot$ & $\cdot$ & $\cdot$ \\
$x_\sigma^1$ & $\cdot$ & 1 & $\cdot$ & 1 & $\cdot$ & $\cdot$ \\
$x_\sigma^2$ & $\cdot$ & $\cdot$ & 1 & 1 & 1 & $\cdot$ \\
\hline
$x_0^0$ & $\cdot$ & $\cdot$ & $\cdot$ & 1 & $\cdot$ & $\cdot$ \\
$x_0^1$ & $\cdot$ & $\cdot$ & $\cdot$ & 1 & 1 & $\cdot$ \\
$x_0^2$ & $\cdot$ & $\cdot$ & $\cdot$ & 1 & 1 & 1 \\
\end{tabular}
\end{table}

For Edit-Mode SFT, the nominal $2N$ layout is $[x_t^{\mathrm{edit}};\mathrm{PAD}^N]$. If $v(j)$ indicates a real non-padding key in the first half,
\begin{equation}
\begin{gathered}
M^{\mathrm{edit}}_{ij}=1\Longleftrightarrow\\
(i<N\land j<N\land v(j))\ \lor\ (i=j\land i\ge N).
\end{gathered}
\label{eq:medit}
\end{equation}
Every real full-canvas token is mutually visible in the first half. The dummy half carries no clean target and retains only a numerical-safety diagonal.

\subsection{Loss support and parameter sharing}

\begin{table*}[t]
\caption{Prediction targets for the two runtime views and MDSAS.}
\label{tab:loss-support}
\centering
\footnotesize
\begin{tabularx}{\textwidth}{>{\raggedright\arraybackslash}p{0.13\textwidth}>{\raggedright\arraybackslash}p{0.22\textwidth}>{\raggedright\arraybackslash}X>{\raggedright\arraybackslash}X}
\toprule
\textbf{Stage} & \textbf{Serialization} & \textbf{Direct loss positions} & \textbf{Conditioning only} \\
\midrule
Action-Mode SFT & $[x_\sigma;x_0]$ & Corrupted tokens in the newly generated Action-Mode segment & Prompt, history, information, uncorrupted target tokens, clean lane, padding \\
Edit-Mode SFT & $[x_t^{\mathrm{edit}};\mathrm{PAD}^N]$ & Controller-owned unresolved masks, using EPM & All other canvas tokens, resolved edit tokens, non-owned plan spans, dummy half \\
MDSAS & $x_{g,k}$ & Generated Action-Mode tokens committed in slice $k$ & Prefix, observations, clean helper blocks, padding \\
\bottomrule
\end{tabularx}
\end{table*}
Table~\ref{tab:loss-support} lists the prediction targets of the two runtime views and of RL. The MDSAS loss covers generated plan and reasoning tokens, tags, search queries, edit requests, and answers. The preceding SFT stage supplies supervision for both action generation and edit reconstruction. The RL stage updates the shared parameters with the trajectory loss and frozen-reference anchor in Equation~\ref{eq:sft_anchor}.

\subsection{Commit-trajectory likelihood}
\label{app:ppo}

\paragraph{Full trajectory objective.}
For trajectory $\tau_g$, the format gate $\gamma(\tau_g)$ equals $1$, $0.5$, or $0$ for clean, soft-invalid, or hard-invalid outputs. The group reward mean and sample standard deviation are $\mu_r$ and $\sigma_r$, with stabilizer $\delta=10^{-4}$. Let $\mathcal I_{g,k}$ contain positions committed in reconstructed slice $k$, and $\Omega_{\mathrm{act}}=\{(g,k,j):j\in\mathcal I_{g,k}\}$ index the Action-Mode loss targets from nonzero-variance trajectory groups. The reward, advantage, likelihood ratio, and clipped objective are
\begin{equation}
\begin{aligned}
    r_g&=\operatorname{F1}(\widehat y_g,\mathcal Y^\star)\gamma(\tau_g),
    \\
    \widehat A_g&=\frac{r_g-\mu_r}{\sigma_r+\delta},\\
    p_{\theta,g,k,j}&=p_\theta(y_{g,j}\mid x_{g,k};M^{\mathrm{act}}),
    \\
    \rho_{g,k,j}&=p_{\theta,g,k,j}/p_{\theta_{\mathrm{old}},g,k,j},\\
    \bar\rho_{g,k,j}&=\operatorname{clip}(\rho_{g,k,j},1-\varepsilon,1+\varepsilon),\\
    \widehat d_{g,k,j}&=\rho_{g,k,j}^{-1}-1+\log\rho_{g,k,j},\\
    \mathcal L_{\mathrm{RL}}^{\mathrm{act}}
    &=-\mathbb E_{\Omega_{\mathrm{act}}}
      \left[\min\!\left(\rho_{g,k,j}\widehat A_g,
      \bar\rho_{g,k,j}\widehat A_g\right)\right]
      \\&\quad+\beta\,\mathbb E_{\Omega_{\mathrm{act}}}[\widehat d_{g,k,j}].
\end{aligned}
\label{eq:action_ppo}
\end{equation}
Here $\theta_{\mathrm{old}}$ is the pre-update policy, $\varepsilon$ is the PPO clip radius, and $\beta$ weights the nonnegative $k_3$ Monte Carlo KL estimator. Zero-variance trajectory groups are omitted from this trajectory loss. Retrieved observations and the token-fixed prefix provide conditioning context, and all Edit-Mode positions are excluded from the loss.

\paragraph{Reconstructing rollout slices.}

During rollout, we record the commit-step index $\zeta_{g,j}$ for each Action-Mode token $y_{g,j}$. These annotations reconstruct each slice $x_{g,k}$ in the two-lane layout of Action-Mode SFT: earlier commitments are visible, positions in $\mathcal I_{g,k}$ are prediction targets, and later positions remain masked. Current and old likelihoods in Equation~\ref{eq:action_ppo} are evaluated on this same slice. The prefix and earlier commitments provide conditioning context. The frozen SFT reference is evaluated on slices from all rollout groups, including zero-variance groups, with the same response-length normalization as the trajectory loss. 

\section{Dynamic Expansion and EPM}
\label{app:epm}

\subsection{Partitions and a minimal example}

This section expands the EPM objective in Equation~\ref{eq:sft_objective}. For component $c$, set $m=m_c$, $n=n_c$, and $y=y^{(c)}$, with $n\ge m$. A latent allocation is a composition
\begin{equation}
\begin{aligned}
\mathcal P_{m,n}=\{\boldsymbol\kappa&=(\kappa_1,\ldots,\kappa_m):\\
&\kappa_i\ge1,\ \textstyle\sum_i\kappa_i=n\}.
\end{aligned}
\end{equation}
Thus the main-text notation is $\mathcal P_c=\mathcal P_{m_c,n_c}$. For compactness, write $p_i(v)=p_\theta(v\mid\widetilde X_{e,c},i;M^{\mathrm{edit}})$ for the label probability at active mask $i$ in component $c$ of edit record $e$. A singleton chunk labels its mask with the target token; a longer chunk labels it with $\chi_{\mathrm{exp}}=\langle\mathrm{EXPAND}\rangle$. For $m=2,n=4$, Table~\ref{tab:epm-example} lists all three valid partitions and their immediate mask labels.

\begin{table}[tbp]
\caption{All EPM partitions for two masks and four target tokens.}
\label{tab:epm-example}
\centering
\footnotesize
\setlength{\tabcolsep}{3pt}
\begin{tabularx}{\linewidth}{@{}>{\centering\arraybackslash}p{0.22\linewidth}>{\centering\arraybackslash}X>{\centering\arraybackslash}X@{}}
\toprule
Partition $(\kappa_1,\kappa_2)$ & Chunk assignment & Immediate mask labels \\
\midrule
$(1,3)$ & $(y_1)\mid(y_2,y_3,y_4)$ & $(y_1,\chi_{\mathrm{exp}})$ \\
$(2,2)$ & $(y_1,y_2)\mid(y_3,y_4)$ & $(\chi_{\mathrm{exp}},\chi_{\mathrm{exp}})$ \\
$(3,1)$ & $(y_1,y_2,y_3)\mid(y_4)$ & $(\chi_{\mathrm{exp}},y_4)$ \\
\bottomrule
\end{tabularx}
\end{table}

The implementation therefore minimizes the uniformly normalized marginal
\begin{equation}
\begin{aligned}
-\log\!\left\{\frac{1}{3}\big[p_1(y_1)p_2(\chi_{\mathrm{exp}})\right.\\
{}+p_1(\chi_{\mathrm{exp}})p_2(\chi_{\mathrm{exp}})\\
\left.{}+p_1(\chi_{\mathrm{exp}})p_2(y_4)\big]\right\}.
\end{aligned}
\label{eq:epm-example}
\end{equation}
Resolved anchors divide a canvas into independent contiguous unresolved groups; EPM is evaluated within each group and the group losses are summed.

\subsection{Dynamic program for partition marginalization}

Let $F_{i,j}$ be the unnormalized log-marginal score for assigning the prefix $(y_1,\ldots,y_j)$ to the first $i$ masks. Define $\operatorname{LSE}(v_1,\ldots,v_d)=\log\sum_{q=1}^d\exp(v_q)$, with the empty log-sum equal to $-\infty$. The boundary conditions are $F_{0,0}=0$ and $F_{0,j>0}=F_{i>0,0}=-\infty$. For $i,j\ge1$,
\begin{equation}
\begin{aligned}
F_{i,j}=\operatorname{LSE}\!\biggl(&
F_{i-1,j-1}+\log p_i(y_j),\\
&\log p_i(\chi_{\mathrm{exp}})+\operatorname{LSE}_{h=0}^{j-2}F_{i-1,h}
\biggr).
\end{aligned}
\label{eq:epm-dp}
\end{equation}
The first branch assigns a singleton ending at $y_j$; the second assigns a chunk of length at least two. Maintaining the inner log-sum-exp as a prefix accumulator gives $O(mn)$ time and space.

Equation~\ref{eq:sft_objective} and the training implementation both use the uniform prior over $|\mathcal P_{m,n}|=\binom{n-1}{m-1}$ paths. For each unresolved group, the implementation evaluates the unnormalized log-sum $F_{m,n}$ from Equation~\ref{eq:epm-dp} and returns
\begin{equation}
\ell_{m,n}^{\mathrm{EPM}}=-F_{m,n}+\log\binom{n-1}{m-1},
\end{equation}
thereby retaining the parameter-independent partition normalizer in the objective.

\subsection{Expansion runtime}

An \texttt{<EXPAND>} prediction replaces one controller-owned mask by two masks with the same owner, increasing the canvas by one token. Each edit invocation permits at most $N_{\mathrm{exp}}^{\max}=32$ expansion events. The operator supports length-preserving and length-growing replacements; deletion and contraction are outside its scope.

Denoising stops when all owned masks are filled or the step limit $T_{\max}=\max\{1,|P|+2N_{\mathrm{exp}}^{\max}+32\}$ is reached, where $|P|$ is the pre-edit plan length in tokens. Expansion is confined to controller-owned Edit-Mode positions.

\section{Teacher Data Construction}
\label{app:teacher_data}

\subsection{Teacher and sampling setup}

Qwen3.6-27B generates trajectories with an indexed plan, retrieval, answer bindings, and replanning. Each retrieval returns three passages truncated to 500 whitespace-delimited words. Collection allows at most six searches, one replan, and two format-repair attempts; a second pass resamples first-pass rejections. The teacher decodes with temperature 0.5, top-$p$ 0.8, top-$k$ 20, and at most 1{,}024 output tokens.

\subsection{Rejection sampling and canonicalization}

We retain trajectories with token-level answer F1 above $0.65$, at least one retrieval, and a valid plan--action structure. The converter maps each teacher trajectory into the two student views. Teacher answer bindings and rewritten questions become \texttt{update} Action records plus full-canvas Edit targets; teacher replacement suffixes become marker-only \texttt{replan} actions plus their corresponding Edit targets. Retrieved \texttt{information} remains conditioning context. Retrieval-limit records teach the agent to answer rather than issue another search.

\begin{table}[tbp]
\caption{Canonical teacher trajectories and derived SFT records.}
\label{tab:record-counts}
\centering
\footnotesize
\begin{tabularx}{\linewidth}{@{}lr>{\raggedright\arraybackslash}X@{}}
\toprule
\textbf{Record family} & \textbf{Count} & \textbf{Breakdown} \\
\midrule
Teacher trajectories & 6{,}486 & --- \\
Action records & 26{,}539 & 12{,}722 search; 6{,}769 update; 6{,}486 answer; 562 replan \\
Edit records & 83{,}519 & 70{,}031 update states; 13{,}488 replan states \\
\bottomrule
\end{tabularx}
\end{table}
Table~\ref{tab:record-counts} counts the canonical trajectories and derived records. Each canonical environment-facing decision yields one Action record. Each update or replan handoff yields multiple deduplicated DreamOn-style full-canvas corruption states (up to 12 per update and 24 per replan), explaining why Edit records outnumber both actions and trajectories.

\section{Full Training Details}
\label{app:training}

\subsection{Optimization configuration}

We run SFT for two epochs and use the epoch-one checkpoint as the reported SFT policy and RL initialization. The training multiset $\mathcal D_{\mathrm{mix}}=\mathcal D_{\mathrm{mix}}^{\mathrm{act}}\uplus\mathcal D_{\mathrm{mix}}^{\mathrm{edit}}$ assigns multiplicity~1 to search, answer, and edit records, multiplicity~1.5 to updates, and multiplicity~3 to replans and search-limit recovery records. The joint SFT objective is
\begin{equation}
\mathcal L_{\mathrm{SFT}}=\frac{1}{|\mathcal D_{\mathrm{mix}}|}
\Bigl[\sum_{r\in\mathcal D_{\mathrm{mix}}^{\mathrm{act}}}\ell_{\mathrm{act}}(r)
+\sum_{e\in\mathcal D_{\mathrm{mix}}^{\mathrm{edit}}}\ell_{\mathrm{edit}}(e)\Bigr].
\end{equation} Table~\ref{tab:training-hparams} summarizes optimization.

\begin{table}[tbp]
\caption{Optimization configuration.}
\label{tab:training-hparams}
\centering
\footnotesize
\begin{tabularx}{\linewidth}{@{}l>{\raggedright\arraybackslash}X>{\raggedright\arraybackslash}X@{}}
\toprule
\textbf{Setting} & \textbf{SFT} & \textbf{MDSAS} \\
\midrule
Initialization & LLaDA2.0-mini MoE merged checkpoint & SFT epoch~1 \\
Optimizer & AdamW & AdamW \\
Weight decay & $0.1$ & $0.01$ \\
Learning rate & peak $2\times10^{-6}$ & constant $1\times10^{-6}$ \\
Schedule & 3\% linear warmup, cosine decay to $10^{-7}$ & no warmup or scheduler \\
Gradient clipping & $5.0$ & $1.0$ \\
\bottomrule
\end{tabularx}
\end{table}
Maximum sequence length is 4{,}096 and diffusion block size is 32. For an Action-Mode target $y$,
\begin{equation}
\sigma\sim\mathcal U(0.3,0.8),\qquad m_i\sim\operatorname{Bernoulli}(\sigma),
\end{equation}
and only target-segment tokens are eligible for corruption. If none is selected, one target position is forced to mask. Edit records receive no additional random corruption: controller ownership defines their EPM support.

\subsection{MDSAS schedule and sampling}
\label{app:anchor_curves}

\begin{figure*}[t]
    \centering
    \includegraphics[width=0.9\textwidth]{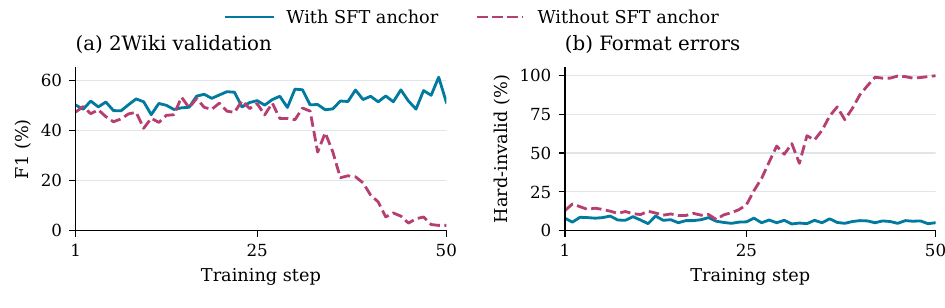}
    \caption{MDSAS and its unanchored variant. (a) Unsmoothed 2Wiki validation F1 and (b) hard-invalid rollout rates over the first 50 training steps.}
    \label{fig:anchor_curves}
\end{figure*}

Training runs for 80 on-policy outer steps, sampling eight trajectories per question. PPO uses a clip of $0.15$ and an old-policy KL coefficient of $0.02$. The fixed SFT anchor has coefficient $\lambda_{\mathrm{SFT}}=0.02$. Each step makes one pass over the newly sampled Action-Mode rows. MDSAS uses a $2N$ representation with 32-token blocks and shrink factor 8, which combines every eight distinct commit steps within a block into one training state. Table~\ref{tab:sampling-hparams} lists the Action-Mode sampling parameters.

The RL training set consists of 4{,}030 examples sampled from the same training dataset used by R1-Searcher.

\subsection{Backbone and protocol comparison}
\label{app:ar_counterpart}

Section~\ref{sec:state_transitions} trains the editable and append protocols on LLaDA2.0-mini and on Ling-mini-2.0. All four agents use the same Qwen3.6-27B teacher trajectories and record multiplicities, two SFT epochs with the epoch-one checkpoint, and the MDSAS schedule above, with RL checkpoints selected on the same 256-question validation set. Under the append protocol, teacher \texttt{update} and \texttt{replan} actions are serialized with the complete revised plan, and at inference the controller adds the generated plan to the history after the request without modifying the earlier plan. Under the editable protocol on Ling-mini-2.0, these actions carry the replacement text for the affected steps, written left to right, and the controller commits it into the planning state as for \textsc{DiffuPlan}, so the training histories of the two editable agents match token for token. Ling-mini-2.0 agents are served with vLLM, and all agents share the retriever and retrieval budget of Section~\ref{sec:setup}. Table~\ref{tab:backbone_protocol_datasets} reports the four agents on each dataset.

\begin{table*}[!t]
\caption{Per-dataset results for the backbone and protocol comparison of Table~\ref{tab:backbone_protocol}. Each dataset reports token F1 and \accl{} (\%); Avg.\ averages the four datasets. Tokens is model-generated tokens per question. Bold marks the best value in each column.}
\label{tab:backbone_protocol_datasets}
\centering
\footnotesize
\setlength{\tabcolsep}{5pt}
\begin{tabular}{l*{11}{c}}
\toprule
& \multicolumn{2}{c}{HotpotQA} & \multicolumn{2}{c}{2Wiki} & \multicolumn{2}{c}{MuSiQue} & \multicolumn{2}{c}{Bamboogle} & \multicolumn{2}{c}{Avg.} & \\
\cmidrule(lr){2-3}\cmidrule(lr){4-5}\cmidrule(lr){6-7}\cmidrule(lr){8-9}\cmidrule(lr){10-11}
\textbf{Planning state} & F1 & \accl & F1 & \accl & F1 & \accl & F1 & \accl & F1 & \accl & Tokens \\
\midrule
\multicolumn{12}{l}{\textit{LLaDA2.0-mini}} \\
Editable & 55.34 & 65.00 & \textbf{55.03} & \textbf{56.00} & 31.72 & 34.60 & \textbf{58.86} & \textbf{65.60} & \textbf{50.24} & \textbf{55.30} & \textbf{338} \\
Append & 54.25 & 64.00 & 51.82 & 53.40 & 28.73 & 34.20 & 56.40 & 57.60 & 47.80 & 52.30 & 411 \\
\midrule
\multicolumn{12}{l}{\textit{Ling-mini-2.0}} \\
Editable & 55.99 & 65.20 & 54.30 & 54.80 & \textbf{33.59} & 35.60 & 56.16 & 59.20 & 50.01 & 53.70 & 357 \\
Append & \textbf{57.48} & \textbf{65.80} & 53.70 & 54.40 & 33.50 & \textbf{35.80} & 53.50 & 54.40 & 49.54 & 52.60 & 425 \\
\bottomrule
\end{tabular}
\end{table*}

\subsection{Edit usage at inference}
\label{app:edit_usage}

Table~\ref{tab:edit_usage} reports how often \textsc{DiffuPlan} edits its plan on the 1{,}625 test questions and how often the executed edits invoke expansion. Edit requests whose markers match no plan span leave the history unchanged and are not counted as executed edits. Table~\ref{tab:edit_usage_paired} pairs \textsc{DiffuPlan} with the LLaDA2.0-mini append agent of Table~\ref{tab:backbone_protocol} on the same multi-hop questions, grouped by the append agent's own edit requests.

\begin{table}[tbp]
\caption{Edit usage of \textsc{DiffuPlan} on the test set. Update and Replan are the percentages of questions with at least one such edit; Upd./q and Rep./q are edits per question; Exp.\ is the percentage of executed edits that invoke expansion.}
\label{tab:edit_usage}
\centering
\footnotesize
\setlength{\tabcolsep}{3.5pt}
\begin{tabular}{lrrrrr}
\toprule
\textbf{Dataset} & Update & Replan & Upd./q & Rep./q & Exp. \\
\midrule
HotpotQA & 71.2 & 9.6 & 1.04 & 0.10 & 88.3 \\
2Wiki & 83.2 & 16.4 & 1.52 & 0.16 & 92.9 \\
MuSiQue & 81.8 & 18.8 & 1.20 & 0.19 & 90.4 \\
Bamboogle & 91.2 & 4.8 & 0.98 & 0.05 & 89.1 \\
\midrule
All & 79.7 & 14.2 & 1.23 & 0.14 & 90.7 \\
\bottomrule
\end{tabular}
\end{table}

\begin{table*}[tbp]
\caption{Paired F1 (\%) and model-generated tokens per question on the 1{,}625 test questions, grouped by the edit requests of the LLaDA2.0-mini append agent. The last two groups are subsets of the second.}
\label{tab:edit_usage_paired}
\centering
\footnotesize
\setlength{\tabcolsep}{3pt}
\begin{tabular}{lrrrrrrr}
\toprule
& & & \multicolumn{3}{c}{F1} & \multicolumn{2}{c}{Tokens} \\
\cmidrule(lr){4-6}\cmidrule(lr){7-8}
\textbf{Append-agent requests} & $n$ & \% & Editable & Append & $\Delta$ & Editable & Append \\
\midrule
No edit request & 223 & 13.7 & 39.40 & 39.20 & +0.20 & 359 & 379 \\
$\geq$1 edit request & 1{,}402 & 86.3 & 49.65 & 46.87 & +2.79 & 335 & 416 \\
\quad $\geq$2 updates & 433 & 26.6 & 55.38 & 50.75 & +4.63 & 383 & 546 \\
\quad $\geq$1 replan & 272 & 16.7 & 30.18 & 23.06 & +7.12 & 433 & 585 \\
\bottomrule
\end{tabular}
\end{table*}

\begin{table}[tbp]
\caption{Action-Mode sampling parameters.}
\label{tab:sampling-hparams}
\centering
\footnotesize
\begin{tabularx}{\linewidth}{@{}l>{\centering\arraybackslash}X>{\centering\arraybackslash}X@{}}
\toprule
& \textbf{Training rollout} & \textbf{Checkpoint selection} \\
\midrule
Block size / denoising steps & 32 / 32 & 32 / 32 \\
Temperature & 0.8 & 0 \\
Top-$p$ / top-$k$ & 0.95 / 0 & 1.0 / 0 \\
Confidence threshold & 0.90 & 0.95 \\
Max new tokens per segment & 256 & 256 \\
\bottomrule
\end{tabularx}
\end{table}
\section{Evaluation Details}
\label{app:experimental_details}

\subsection{Retriever and corpus}

All retrieval-based methods use \texttt{intfloat/\allowbreak e5-base-v2}~\citep{e5} with mean pooling over the Wiki18 corpus and FAISS inner-product search. Queries are truncated to 256 encoder tokens.

\subsection{Prompts, answer extraction, and scoring}

Direct, CoT, and Standard RAG use concise answer-tag prompts; other systems use their released or paper-specified workflows and native answer markers. A method-specific parser extracts the terminal answer. The common scorer lowercases text, removes punctuation and English articles, collapses whitespace, and takes the maximum score over reference aliases. EM compares normalized strings, token F1 measures token overlap, and SubEM tests whether a normalized reference appears in the prediction.

\subsection{LLM-as-Judge}
\label{app:llm_judge}

Lexical metrics penalize correct answers phrased differently from the references and reward responses that contain a reference without committing to it. We therefore also score every extracted answer with LLM-as-Judge, using Qwen3.5-9B~\citep{qwen35} as the judge model with thinking disabled and greedy decoding. The judge model receives the question, the reference aliases, and the extracted answer, truncated to 300 words, and returns a binary verdict. Its instructions treat aliases, abbreviations, name order, formatting differences, and additional correct detail as equivalent, and mark as incorrect an answer that is vaguer than every reference, names a different entity, lists several candidates, declares the answer unknown, contradicts a reference, or is empty; yes/no questions are compared by polarity only. Pooled over the twelve systems and 1{,}625 questions, the verdicts agree with EM on 86.5\% of answers, with SubEM on 92.7\%, and with nonzero F1 on 88.1\%. Table~\ref{tab:system_comparison} reports LLM-as-Judge accuracy beside F1 for every dataset, and Table~\ref{tab:backbone_protocol_datasets} for the backbone and protocol comparison.

\subsection{Format validity and reward gating}

Format validity measures the fraction of trajectories with correct output structure, independently of answer correctness. The checks require an initial plan, balanced control tags, well-formed closing tags, a non-empty terminal answer without control markup, and valid reasoning--search transitions. Empty searches, duplicate answer blocks, and model-generated information blocks are invalid. Both SFT diagnostics in Figure~\ref{fig:edit_learning}(a,b) use this same definition.

For the RL reward gate in Equation~\ref{eq:action_ppo}, malformed control structures and missing or malformed terminal answers are hard-invalid. Edit-marker payloads are also checked for valid target syntax. Recoverable deviations, including an empty search or a missing reasoning transition between searches, are soft-invalid. The hard-invalid rollout rate in Figure~\ref{fig:anchor_curves} uses these training-time checks.

\subsection{Answer-output metrics}
\label{app:answer_output}

EM and SubEM complement token F1 by measuring exact answers and the presence of a reference answer within the response; LLM-as-Judge (Appendix~\ref{app:llm_judge}) scores semantic correctness (Table~\ref{tab:answer_output_diagnostic}).

\begin{table}[tbp]
\caption{Answer metrics averaged over the four datasets (\%).}
\label{tab:answer_output_diagnostic}
\centering
\footnotesize
\setlength{\tabcolsep}{4pt}
\begin{tabular}{lcccc}
\toprule
\textbf{System} & \textbf{EM} & \textbf{F1} & \textbf{SubEM} & \textbf{\accl} \\
\midrule
Search-o1 & 29.45 & 39.78 & 39.50 & 46.25 \\
Search-R1 & 27.50 & 35.33 & 33.90 & 37.75 \\
R1-Searcher & 38.40 & 48.71 & 46.55 & 52.60 \\
DLLM-Searcher & 0.05 & 7.25 & 36.25 & 36.65 \\
\textsc{DiffuPlan} & 38.15 & 50.24 & 48.45 & 55.30 \\
\bottomrule
\end{tabular}
\end{table}

\section{Qualitative Examples}
\label{app:qualitative}

The following traces retain every model action. Each \texttt{Controller commit} block shows a snapshot of the same plan after it has been edited in place. Only the retrieved passages inside \texttt{<information>} are abridged.
\lstset{aboveskip=1pt,belowskip=1pt,breakatwhitespace=true}

\subsection{Successful update}
\label{app:qualitative_update}

For the question ``Who wrote the 1897 novel about a vampire which was later turned into a 1931 film featuring B\'ela Ferenc Dezs\H{o} Blask\'o?'', the agent performs two successive updates:
\begin{lstlisting}
Question: Who wrote the 1897 novel about a vampire which was later turned into a 1931 film featuring Bela Ferenc Dezso Blasko?
Student:
<plan>
#Q_1: Who is Bela Ferenc Dezso Blasko?
#Q_2: What 1931 film featuring #A_1 was based on an 1897 vampire novel?
#Q_3: Who wrote the 1897 novel #A_2?
</plan>
<think>I need to identify the actor Bela Ferenc Dezso Blasko to find the 1931 film he was in.</think>
<search>Bela Ferenc Dezso Blasko actor</search>
Environment (abridged):
<information>Bela Ferenc Dezso Blasko is better known as Bela Lugosi, the actor famous for portraying Count Dracula in the 1931 film.</information>
Student:
<think>The retrieved evidence states that Bela Ferenc Dezso Blasko is better known as Bela Lugosi.</think>
<update>#A_1</update>
Controller commit:
<plan>
#Q_1: Who is Bela Ferenc Dezso Blasko?
#Q_2: What 1931 film featuring Bela Lugosi was based on an 1897 vampire novel?
#Q_3: Who wrote the 1897 novel #A_2?
</plan>
Student:
<think>I need to identify the 1931 film featuring Bela Lugosi that was based on an 1897 vampire novel.</think>
<search>1931 film featuring Bela Lugosi based on 1897 vampire novel</search>
Environment (abridged):
<information>Dracula is a 1931 vampire film starring Bela Lugosi and based on the novel Dracula by Bram Stoker.</information>
Student:
<think>The retrieved information confirms that the 1931 film Dracula is based on the novel Dracula by Bram Stoker.</think>
<update>#A_2</update>
Controller commit:
<plan>
#Q_1: Who is Bela Ferenc Dezso Blasko?
#Q_2: What 1931 film featuring Bela Lugosi was based on an 1897 vampire novel?
#Q_3: Who wrote the 1897 novel Dracula?
</plan>
Student:
<think>I need to find the author of the 1897 novel Dracula.</think>
<search>author of 1897 novel Dracula</search>
Environment (abridged):
<information>Dracula is an 1897 Gothic horror novel by Irish author Bram Stoker.</information>
Student:
<think>The search results confirm that Bram Stoker is the author of the 1897 novel Dracula.</think>
<answer>Bram Stoker</answer>
\end{lstlisting}
The first commit binds the actor's stage name into \texttt{\#Q\_2}; the second binds the film title into \texttt{\#Q\_3}. The final answer matches the reference, Bram Stoker.

\subsection{Successful replan}
\label{app:qualitative_replan}

For the question ``Which 2009 film features an actor that also starred in \emph{Bad Samaritan}?'', the agent first updates the cast binding and then replans after an incomplete retrieval:
\begin{lstlisting}
Question: Which 2009 film features an actor that also starred in Bad Samaritan?
Student:
<plan>
#Q_1: Who are the main actors in Bad Samaritan?
#Q_2: Which 2009 film features #A_1?
</plan>
<think>I need to identify the cast of the film Bad Samaritan to find the actor in question.</think>
<search>Bad Samaritan film cast</search>
Environment (abridged):
<information>Bad Samaritan stars David Tennant and Robert Sheehan, with Carlito Olivero, Kerry Condon, and Jacqueline Byers.</information>
Student:
<think>The observation provides the evidence needed for the current step.</think>
<update>#A_1</update>
Controller commit:
<plan>
#Q_1: Who are the main actors in Bad Samaritan?
#Q_2: Which 2009 film features David Tennant or Robert Sheehan?
</plan>
Student:
<think>I will search for a 2009 film featuring either of these actors to find the answer.</think>
<search>2009 film starring David Tennant or Robert Sheehan</search>
Environment (abridged):
<information>Robert Sheehan appeared with Rupert Grint in an unnamed 2009 film set in Belfast; the retrieved text omits its title.</information>
Student:
<think>The observation provides the plot and co-star (Rupert Grint) for Robert Sheehan's 2009 film but does not explicitly state the title. I will replan to search for the title using these constraints.</think>
<replan>#Q_2</replan>
Controller commit:
<plan>
#Q_1: Who are the main actors in Bad Samaritan?
#Q_2: What is the title of the 2009 film starring Robert Sheehan and Rupert Grint set in Belfast?
</plan>
Student:
<think>I will search for the title of the 2009 film starring Robert Sheehan and Rupert Grint.</think>
<search>2009 film starring Robert Sheehan and Rupert Grint set in Belfast</search>
Environment (abridged):
<information>Cherrybomb is a 2009 British drama starring Rupert Grint and Robert Sheehan and filmed in Belfast.</information>
Student:
<think>The observation provides the evidence needed for the current step.</think>
<answer>Cherrybomb</answer>
\end{lstlisting}
The replan preserves the resolved cast step and rewrites only \texttt{\#Q\_2}. The final answer matches the reference, Cherrybomb.

\end{document}